%% file: iclr2025_conference.tex
\documentclass{article} 
\usepackage{iclr2025_conference,times}

\input{math_commands.tex}

\usepackage[utf8]{inputenc} 
\usepackage[T1]{fontenc}    
\usepackage[colorlinks=true, citecolor=blue, linkcolor=blue, urlcolor=blue]{hyperref}
\usepackage{url}            
\usepackage{booktabs}       
\usepackage{amsfonts}       
\usepackage{nicefrac}       
\usepackage{microtype}      
\usepackage{listings}
\usepackage{xcolor}         
\usepackage{algorithm}
\usepackage{bm}
\usepackage{svg}
\usepackage{float}
\usepackage{tikz}
\usepackage{subcaption}
\usepackage{caption}
\usepackage{algpseudocode}
\usepackage{comment}
\usepackage{mdframed}
\usepackage{array}
\usepackage{arydshln}
\usepackage{amsmath, amssymb, amsthm}
\usepackage{siunitx}
\usepackage{algorithm}
\usepackage{tikz}
\usepackage{multirow}
\usepackage{xcolor}
\usepackage{xcolor}
\definecolor{ForestGreen}{rgb}{0.13, 0.55, 0.13}

\usepackage{algpseudocode}
\DeclareMathOperator{\logits}{logits}

\title{Fast and Simplex: 2-Simplicial Attention in Triton}

\author{%
  Aurko Roy \\
  Meta \\
  Menlo Park, CA \\
  \texttt{roy.aurko@gmail.com}
  \And
  Timothy Chou \\
  Meta \\
  Menlo Park, CA \\
  \texttt{timchou@meta.com}
  \And 
  Sai Surya Duvvuri\thanks{Work done during an internship at Meta} \\
  Department of Computer Science \\
  University of Texas at Austin \\
  \texttt{saisurya@cs.utexas.edu}
  \And 
  Sijia Chen \\
  Meta \\
  Menlo Park, CA \\
  \texttt{sijiac@meta.com}
  \And 
  Jiecao Yu \\
  Meta \\
  Menlo Park, CA \\
  \texttt{jiecaoyu@meta.com}
  \And 
  Xiaodong Wang \\
  Meta \\
  Menlo Park, CA \\
  \texttt{xdwang@meta.com}
  \And 
  Manzil Zaheer \\
  Meta \\
  Menlo Park, CA \\
  \texttt{manzilzaheer@meta.com}
  \And
  Rohan Anil\thanks{Work done while at Meta} \\
  San Francisco, CA \\
  \texttt{rohan.anil@gmail.com}
}

\renewcommand{\a}{\mathbf{a}}
\renewcommand{\b}{\mathbf{b}}
\renewcommand{\c}{\mathbf{c}}
\renewcommand{\R}{\mathbb{R}}
\iclrfinalcopy 
\begin{document}

\maketitle

\begin{abstract}
Recent work has shown that training loss scales as a power law with both model size and the 
number of tokens, and that achieving compute-optimal models requires scaling model size and 
token count together. However, these scaling laws assume an infinite supply of data and 
apply primarily in compute-bound settings. As modern large language models increasingly rely 
on massive internet-scale datasets, the assumption that they are compute-bound is becoming 
less valid. This shift highlights the need for architectures that prioritize token 
efficiency.

In this work, we investigate the use of the 2-simplicial Transformer, an architecture that 
generalizes standard dot-product attention to trilinear functions through an efficient 
Triton kernel implementation. We demonstrate that the 2-simplicial Transformer achieves 
better token efficiency than standard Transformers: for a fixed token budget, similarly 
sized models outperform their dot-product counterparts on tasks involving mathematics, 
coding, reasoning, and logic. We quantify these gains by demonstrating
that $2$-simplicial attention changes the exponent 
in the scaling laws for knowledge and reasoning tasks compared to dot product attention.
\end{abstract}

\section{Introduction}\label{sec:intro}
Large language models (LLMs) based on the Transformer architecture \citep{vaswani2017attention} have become foundational to many state-of-the-art artificial intelligence systems, including 
GPT-3 \citep{brown2020language}, GPT-4 \citep{achiam2023gpt}, Gemini \citep{team2023gemini}, and Llama \citep{touvron2023llama}. The remarkable progress in scaling these models has been 
guided by neural scaling laws \citep{hestness2017deep, kaplan2020scaling, hoffmann2022training}, which empirically establish a power-law relationship between training loss, the number of 
model parameters, and the volume of training data.

A key insight from this body of work is that optimal model performance is achieved not simply by increasing model size, but by scaling both the number of parameters and the amount of 
training data in tandem. Notably, \cite{hoffmann2022training} demonstrate that compute-optimal models require a balanced scaling approach. Their findings show that the Chinchilla model, with 
70 billion parameters, outperforms the much larger Gopher model (280 billion parameters) by being trained on four times as much data. This result underscores the importance of data scaling 
alongside model scaling for achieving superior performance in large language models.

As artificial intelligence (AI) continues to advance, a significant emerging challenge is the availability of sufficiently high-quality tokens. 
As we approach this critical juncture, it becomes imperative to explore novel methods and architectures that can scale more efficiently than traditional
Transformers under a limited token budget. However, most architectural and optimizer improvements merely shift the error but do not meaningfully 
change the exponent of the power law~\citep{tweet_katie}. The work of \cite{kaplan2020scaling, shen2024scaling} showed that most architectural 
modifications do not change the exponent, while \cite{hestness2017deep} show a similar result for optimizers. The only positive result has been on
data due to the works of \cite{sorscher2022beyond,bahri2024explaining,brandfonbrener2024loss} who show that changing the data
distribution can affect the exponent in the scaling laws.

In this context we revisit an old work \cite{clift2019logic} which
generalizes the dot product attention of
Transformers to trilinear forms as the $2$-simplicial Transformer. 
We explore generalizations of RoPE~\citep{su2024roformer} to trilinear functions and
present a rotation invariant trilinear form that we prove is as 
expressive as $2$-simplicial
attention.
We further show that the $2$-simplicial Transformer 
scales better than the Transformer under a 
limited token budget: for a fixed number of tokens, a similar sized
$2$-simplicial Transformer out-performs the Transformer on math, coding and
reasoning tasks. Furthermore, 
our experiments also reveal that the $2$-simplicial Transformer
has a more favorable scaling exponent corresponding to the number of parameters than the 
Transformer~\citep{vaswani2017attention}.
This suggests that, unlike Chinchilla scaling~\citep{hoffmann2022training}, it is
possible to increase tokens at a slower rate than the parameters for the 
$2$-simplicial Transformer. Our findings imply that, when operating under token constraints, the 2-simplicial Transformer can more effectively approach the irreducible 
entropy of natural language compared to dot product attention Transformers.

\section{Related work}
Several generalizations of attention have been proposed since the seminal
work of \cite{vaswani2017attention}. A line of work that started immediately after
was to reduce the quadratic complexity of attention with sequence length. 
In particular, the work of \cite{parmar2018image} proposed local attention in the
context of image generation and several other works subsequently used it in conjunction
with other methods for language modeling~\citep{zaheer2020big, roy2021efficient}. Other 
work has proposed doing away with softmax attention altogether - e.g.,
\cite{katharopoulos2020transformers} show that replacing the softmax with an  
exponential without normalization leads to linear time Transformers using the 
associativity of matrix multiplication. Other linear time attention work are
state space models such as Mamba \citep{gu2023mamba}; however these linear time
attention methods have received less widespread adoption due to their worse
quality compared to Transformers in practice. According to Allen (2025), the key factor 
contributing to Mamba's success in practical applications is the utilization of the 
$\operatorname{conv1d}$ operator;
see also \cite{so2021searching} and \cite{roy2022n} for similar proposals to the
Transformer architecture.

The other end of the spectrum is going from quadratic to higher order attention.
The first work in this direction to the best of our knowledge
was $2$-simplicial attention proposed by 
\cite{clift2019logic} which showed
that it is a good proxy for logical problems in the context of deep reinforcement
learning. A similar generalization of Transformers was proposed in 
\cite{bergen2021systematic} which proposed the \emph{Edge Transformer} where
the authors proposed \emph{triangular attention}.
The AlphaFold~\citep{jumper2021highly}
paper also used an attention mechanism similar to the \emph{Edge Transformer} which
the authors called \emph{triangle self-attention} induced by the $2D$ geometry
of proteins. Higher order interactions were also explored in \cite{wang2021dcn}
in the context of recommender systems.
Recent work by \cite{sanford2023representational} 
shows that the class of problems solved by an $n$-layer 
$2$-simplicial Transformer is strictly larger than the 
class of problems solved by dot product
attention Transformers. In particular, the authors define a class of problems
referred to as $\texttt{Match3}$ and show that dot product attention requires
exponentially many layers in the sequence length to solve this task. 
Follow up work by \cite{kozachinskiy2025strassen}
propose a scalable approximation to $2$-simplicial attention and prove 
lowerbounds between Strassen attention and dot product attention on tasks
that require more complex reasoning using VC dimension
\citep{vapnik1968uniform} arguments. 

Also related is work on looping Transformer layers \citep{dehghani2018universal}
as in Universal Transformers; see also \cite{yang2023looped, saunshi2025reasoning} 
for a more recent treatment of the same idea. Both higher order attention and looping 
serve a similar purpose: compute a more expressive function per parameter. It has been
established in these works that looped Transformers are better at logical reasoning
tasks. A key challenge in scaling looped Transformers to larger models is their 
trainability. Specifically, looping $k$
times increases the model depth by a factor of 
$k$, which can significantly exacerbate the difficulties associated with training 
deeper models. As a result, it remains unclear how well large looped Transformers can 
be trained, and further research is needed to address this concern.

\paragraph{Notation.} We use small and bold letters to denote vectors, capital letters to denote matrices and tensors and small letters to denote scalars. We denote $\inangle{\a,\b}$ to denote dot product between two vectors $\a$ and $\b$. Similarly, the trilinear dot product is denoted as follows:$\inangle{\a,\b,\c} =  \sum_{i=1}^d \inangle{\a_i,\b_i,\c_i}$.
We use $@$ to highlight a matrix multiplication, for e.g., $(AB)@C$, for matrices $A$, $B$, $C$. To denote array slicing, we use $\a[l:l+m] = (a_l,\ldots,a_{l+m-1})$ with zero-based indexing. Some tensor operations are described using Einstein summation notation  as used in the Numpy library \citep{harris2020array}. We use $FLOPs$ to denote floating point operations.  Column stacking of arrays are denoted by $[\a,\b,\c]$. We use $\det$ to denote determinant of a square matrix.

\section{Overview of neural scaling laws}\label{sec:overview-scaling-laws}
In this section we provide a brief overview of neural scaling laws as introduced
in \cite{kaplan2020scaling}. We will adopt the approach outlined by \cite{hoffmann2022training},
which proposes that the loss $L(N, D)$ decays as a power law in the total number of model 
parameters $N$ and the number of tokens $D$:
\begin{equation}\label{eq:scaling-laws}
L(N, D) = E + \frac{A}{N^\alpha} + \frac{B}{D^\beta}.
\end{equation}
The first term $E$ is often described as the \emph{irreducible loss} which corresponds
to the entropy of natural text. The second term captures the fact that a model with 
$N$ parameters underperforms this ideal generative process. The third term corresponds
to the fact that we train on only a finite sample of the data and do not train the model
to convergence. Theoretically, as $N \rightarrow \infty$ and $D \rightarrow \infty$ a
large language model should approach the irreducible loss $E$ of the underlying text
distribution.

For a given compute budget $C$ where $FLOPs(N, D) = C$, 
one can express the optimal number of parameters as $N_{opt} \propto C^a$ and 
the optimal dataset size as $D_{opt} \propto C^b$. The authors of \cite{hoffmann2022training} 
perform several experiments and fit parametric functions to the loss to estimate the exponents
$a$ and $b$: multiple different approaches confirm that roughly $a \sim 0.49$ while $b \sim 0.5$.
This leads to the central thesis of \cite{hoffmann2022training}: one must
scale the number of tokens proportionally to the model size.

However, as discussed in Section~\ref{sec:intro}, the quantity of sufficiently
high-quality tokens is an emerging bottleneck in pre-training scaling, necessitating
an exploration of alternative training algorithms and architectures. On the other hand
recent studies have shown that most modeling and optimization techniques proposed in the literature merely
shift the error (offset $E$) and do not fundamentally change the exponent in the power law. We
refer the readers to this excellent discussion in \cite{tweet_katie}. 

\section{The $2$-simplicial Transformer}
\begin{figure}[htbp]
    \centering
    \begin{subfigure}[b]{0.45\textwidth}
        \centering
        \begin{tikzpicture}
            \coordinate (i) at (0, 0);
            \coordinate (j) at (3, 0);
            \filldraw (i) circle (2pt);
            \filldraw (j) circle (2pt);
            \draw (i) -- (j) -- cycle;
            \node[below left] at (i) {$i$};
            \node[below right] at (j) {$j$};
        \end{tikzpicture}
        \caption{1-simplex between two nodes $i, j$}
        \label{fig:triangle1}
    \end{subfigure}
    \hfill
    \begin{subfigure}[b]{0.45\textwidth}
        \centering
        \begin{tikzpicture}
            \coordinate (i) at (0, 0);
            \coordinate (j) at (3, 0);
            \coordinate (k) at (1.5, 2.5);
            \filldraw (i) circle (2pt);
            \filldraw (j) circle (2pt);
            \filldraw (k) circle (2pt);
            \draw (i) -- (j) -- (k) -- cycle;
            \node[below left] at (i) {$i$};
            \node[below right] at (j) {$j$};
            \node[above] at (k) {$k$};
        \end{tikzpicture}
        \caption{2-simplex between three nodes $i, j, k$}
        \label{fig:triangle2}
    \end{subfigure}
    \caption{Geometry of dot product attention and $2$-simplical attention.}
    \label{fig:twotriangles}
\end{figure}
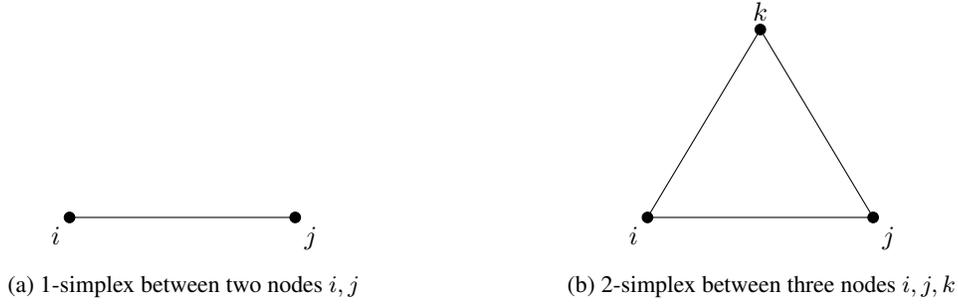

The $2$-simplicial Transformer was introduced in \cite{clift2019logic} where the
authors
extended the dot product attention from bilinear to trilinear forms,
or equivalently from the 1-simplex to the 2-simplex.
Let us recall
the attention mechanism in a standard Transformer \citep{vaswani2017attention}. Given
a sequence $X \in \R^{n \times d}$ we have three projection matrices 
$W_Q, W_K, W_V \in \R^{d \times d}$ which we refer to as the query, key and
value projections respectively. These projection matrices are used to infer the 
query $Q = X W_Q$, key $K = X W_K$ and value $V = X W_V$ respectively. This is then used
to construct the\textit{ attention logits}:
\begin{equation}
A = {QK^\top}/{\sqrt{d}} \in \R^{n\times n},
\end{equation}
where each entry is a dot product 
$A_{ij} = \inangle{\vq_i,\vk_j}/\sqrt{d}$ which are both entries in $\R^d$ 
. 
The attention scores (logits) are then transformed
into probability weights by using a row-wise softmax operation:
\begin{equation}
S_{ij} = {\exp{(A_{ij}})}/{\sum^n_{j=1} \exp{(A_{ij}})}.
\end{equation}
The final output of the attention layer is then a linear combination of the values
according to these attention scores:
\begin{equation}
\tilde{\vv}_i = \sum^n_{j=1} A_{ij} \vv_j
\end{equation}

The $2$-simplicial Transformer paper \cite{clift2019logic} generalizes this to 
trilinear products where we have two additional key and value projection
matrices $W_{K'}$ and $W_{V'}$, which give us $K' = X W_{K'}$ and $V' = X W_{V'}$.
The attention logits for $2$-simplicial Transformer are then given by the trilinear product
between $Q$, $K$ and $K'$, resulting in the following third-order tensor:
\begin{equation}\label{eq:3d-attention}
A_{ijk}^{(\text{2s})} = \frac{\inangle{\vq_i, \vk_j, \vk'_k}}{\sqrt{d}} =
\frac{1}{\sqrt{d}}\sum^d_{l=1} Q_{il} K_{jl} K'_{kl},
\end{equation}
so that the attention tensor becomes:
\begin{equation}
\label{eq:softmax}
S_{ijk}^{(\text{2s})} = {\exp(A_{ijk}^{(\text{2s})})}/{\sum_{j,k}\exp (A_{ijk}^{(\text{2s})})},
\end{equation}
with the final output of the attention operation being defined as
\begin{equation}
\label{eq:attenval}
\tilde{\vv}^{(\text{2s})}(i) = \sum^n_{j, k = 1} S_{ijk}^{(\text{2s})} \left( \vv_j \circ \vv_k' 
\right),
\end{equation}
where $\vv_j \circ \vv'_k$ represents the element wise Hadamard product between two vectors
in $\R^d$. The pseudo-code for $2$-simplicial attention is depicted in
Algorithm~\ref{alg:pseudo-code-simplicial}. Note that Equation~\ref{eq:3d-attention}
does not incorporate any position encoding such as RoPE~\citep{su2024roformer}; we 
discuss this in the next section.

\begin{algorithm}
  \caption{Pseudocode for the forward pass of 2-simplicial attention}
  \label{alg:pseudo-code-simplicial}
  \begin{algorithmic}[1]
    \Procedure{2-simplicial attention}{$Q$, $K$, $V$, $K'$, $V'$}
      \State $\logits \gets \mathrm{einsum(``btnh,bsnh,brnh\rightarrow bntsr", Q, K, K')}$ \label{alg:compute-logits}
      \State $\operatorname{attention} \gets \operatorname{softmax}(\logits + \operatorname{causal-mask}, \operatorname{axis}=[-1, -2])$
      \State $\operatorname{output} \gets \mathrm{einsum(``bntsr,bsnh,brnh\rightarrow btnh", \operatorname{attention}, V, V')}$ 
      \State \Return $\operatorname{output}$
    \EndProcedure
  \end{algorithmic}
\end{algorithm}

\section{Determinant based Trilinear Forms}
RoPE~\citep{su2024roformer} was proposed as a way to capture the positional 
information in a sequence for Transformer language models. RoPE applies a position
dependent rotation to the queries $\vq_i$ and the key $\vk_j$ so that the dot product
$\inangle{\vq_i, \vk_j}$ is a function of the relative distance $i-j$. In particular,
note that the dot product is 
invariant to orthogonal transformations $R\in \R^{d\times d}$:
\begin{align*}
    \inangle{\vq_i,\vk_j} = \inangle{R\vq_i,R\vk_j}.
\end{align*}
This is important for RoPE to work as for a query $\vq_i$ and key $\vk_i$ at the same
position $i$, we expect its dot product to be unchanged by the application of
position based rotations: $\inangle{\vq_i, \vk_i} = \inangle{R\vq_i, R\vk_i}$.

Note that the trilinear form defined in Equation~\ref{eq:3d-attention} is not invariant
to rotation and the application of the same rotation to $\vq_i$, $\vk_i$ and $\vk'_i$
no longer preserves the inner product: $\inangle{\vq_i, \vk_i, \vk'_i} = 
\sum_{l=1}^d \vq_{il}\vk_{il}\vk'_{il}\neq 
\inangle{R\vq_i, R\vk_i, R\vk_i'}$. Therefore, to generalize RoPE to $2$-simplicial
attention, it is important to explore alternative bilinear and trilinear forms
that are rotation invariant.

We note that the following functions are also invariant to rotations: 
\begin{align}
\hat{f}_2(\a,\b)&=\text{det}{\begin{pmatrix} a_{1} & a_2\\
    b_1 & b_2\end{pmatrix}} = a_1b_2 - a_2b_1,\nonumber\\
\hat{f}_3(\a,\b,\c)&=\text{det}{\begin{pmatrix} a_{1} & a_2& a_3\\
    b_1 & b_2 & b_3\\
    c_1 & c_2 & c_3\end{pmatrix}},\nonumber\\
    &= a_1b_2 c_3 + a_2b_3c_1 + a_3b_1c_2  - a_1b_3c_2 - a_2b_1c_3 - a_3b_2c_1 \nonumber\\
    &= \langle (a_1,a_2,a_3), (b_2,b_3,b_1), (c_3, c_1, c_2)  \rangle - \langle (a_1,a_2,a_3), (b_3, b_1, b_2), (c_2, c_3, c_1) \rangle, \label{eq:det}
\end{align}
the rearrangement in the last equality is popularly called Sarrus rule \citep{strang2022introduction}.  Here, $\hat{f}_2$ is a bilinear form in $\a=(a_1,a_2)$ and $\b=(b_1,b_2)$ and $\hat{f}_3$ is a trilinear form in $\a=(a_1,a_2,a_3),\ \b=(b_1,b_2,b_3),\ \c=(c_1,c_2,c_3)$. Geometrically, $\inabs{\hat{f}_2(\a,\b)}$ measures the area of the parallelogram spanned by $\a$ and $\b$, and similarly, $\inabs{\hat{f}_2(\a,\b,\c)}$ measures the volume of the parallelotope spanned by $\a,\ \b$ and $\c$. We use the signed determinant operation $\hat{f}_3$ to compute $A^{(\text{det})}\in \R^{n\times n\times n}$. For any vector $\vq$, let $\vq^{(l)} = \vq = \vq[3(l-1):3l]$ be its $l$th chunk of size 3. The logits are defined as:
\begin{align}
\label{eq:logits}
    A_{i j_1 j_2}^{(\text{det})} = \sum_{l=1}^{p} \det([\vq_i^{(l)}, \vk_{j_1}^{(l)}, {\vk'_{j_2}}^{(l)}]).
\end{align}
Since Equation~\ref{eq:det} has $2$ dot product terms due to Sarrus rule, it would modify
Algorithm~\ref{alg:pseudo-code-simplicial} to use $2$ einsums instead of $1$ in line~\ref{alg:compute-logits}.
The final attention weights $S$ are computed by applying a softmax function on the
logits above, similar to Equation~\ref{eq:softmax}. The output for token $i$ is then the weighted 
sum of value vectors as in Equation~\ref{eq:attenval}.

\begin{theorem}
\label{thm:det}
For any input size $n$ and input range $m=n^{O(1)}$, there exists a transformer architecture with a single head of attention with logits computed as in (\ref{eq:logits}), with attention head dimension $d=7$, such that for all $X \in [M]^N$, the transformer's output for element $x_i$ is 1 if $\exists j_1, j_2$ s.t. $x_i + x_{j_1} + x_{j_2} = 0 \pmod M$, and 0 otherwise.
\end{theorem}
We provide the proof in Appendix~\ref{sec:proof-rotation-invariance}. Since the sum-of-determinants
trilinear function of Equation~\ref{eq:logits} involves $6$ terms compared to the simpler 
trilinear form of Equation~\ref{eq:3d-attention}, in the following sections where we compute the 
backwards function for $2$-simplicial attention, we will use the simpler trilinear form of
Equation~\ref{eq:3d-attention} without loss of generality.

\section{Model design}
Since $2$-simplicial attention scales as $\bigO(n^3)$ in the sequence length $n$, 
it is impractical to apply it over the entire sequence. Instead, we parametrize it as $\bigO(n \times w_1 
\times w_2)$, where $w_1$ and $w_2$ define the dimensions of a sliding window over the sequence. Each query vector $Q_i$ attends to a localized region of $w_1$ $K$ keys and $w_2$ $K'$ keys, thereby reducing the computational burden. We systematically evaluate various configurations of $w_1$ and $w_2$ to identify optimal trade-offs between computational efficiency and model performance (see Table~\ref{tab:w1w2-sweep}). 

For causal dot product attention, the complexity for a sequence of length $n$ is given by:
\begin{equation*}
O(A) = \frac{1}{2} \cdot 2 \cdot 2n^2 = 2n^2,
\end{equation*}
where $n$ is the sequence length. This involves two matrix multiplications: one for $Q @ K$, one for $P @ V$, each requiring two floating-point operations per element. The causal mask allows us to skip $\frac{1}{2}$ of these computations. \\ \\
In contrast, the complexity of $2$-simplical attention, parameterized
by $w_1$ and $w_2$, is expressed as:
\begin{equation*}
O(A^{(2s)}) = 3 \cdot 2n w_1 w_2 = 6n w_1w_2
\end{equation*}
This increase in complexity arises from the trilinear einsum operation, which necessitates an additional multiplication compared to standard dot product attention.

\begin{table}[H]
\centering
\small
\begin{tabular}{lccc}
\toprule
$w_1 \times w_2$ & $w_1$ & $w_2$ & \textbf{Latency (ms)}  \\
\midrule
$32k$ & $1024$ & $32$ & $104.1$ ms\\
$32k$ & $512$ & $64$ & $110.7$ ms \\
$16k$ & $128$ & $128$ & $59.2$ ms\\
$16k$ & $256$ & $64$  & $55.8$ ms\\
$16k$ & $512$ & $32$  & $55.1$ ms \\
$16k$ & $1024$ & $16$ & $55.1$ ms \\
$8k$ & $256$ & $32$ & $28.3$ ms \\

\bottomrule
\end{tabular}
\caption{Latency for different combinations of $w_1$, $w_2$}
\label{tab:w1w2-sweep}
\end{table}

We choose a window size of (512, 32), balancing latency and quality. With this configuration, the computational complexity of
$2$-simplical attention is comparable to dot product attention at $48k$ context length. 

A naive sliding window $2$-simplicial attention implementation has each $Q_i$ vector attending 
to $w_1 + w_2 - 1$ different $KK'$ vectors, as illustrated in Figure~\ref{fig:2-simplical-tiling}. Thus, tiling queries $Q$
like in flash attention leads to poor
compute throughput. Inspired by Native Sparse 
Attention~\citep{yuan2025native}, we adopt a model architecture leveraging a high Grouped Query Attention 
GQA~\citep{ainslie2023gqa} ratio of $64$ . This approach enabled efficient tiling along query heads, ensuring dense computation and eliminating the need for costly element-wise masking.

\section{Kernel Optimization}
\begin{figure}{}
\centering
\includegraphics[width=1.0\textwidth]{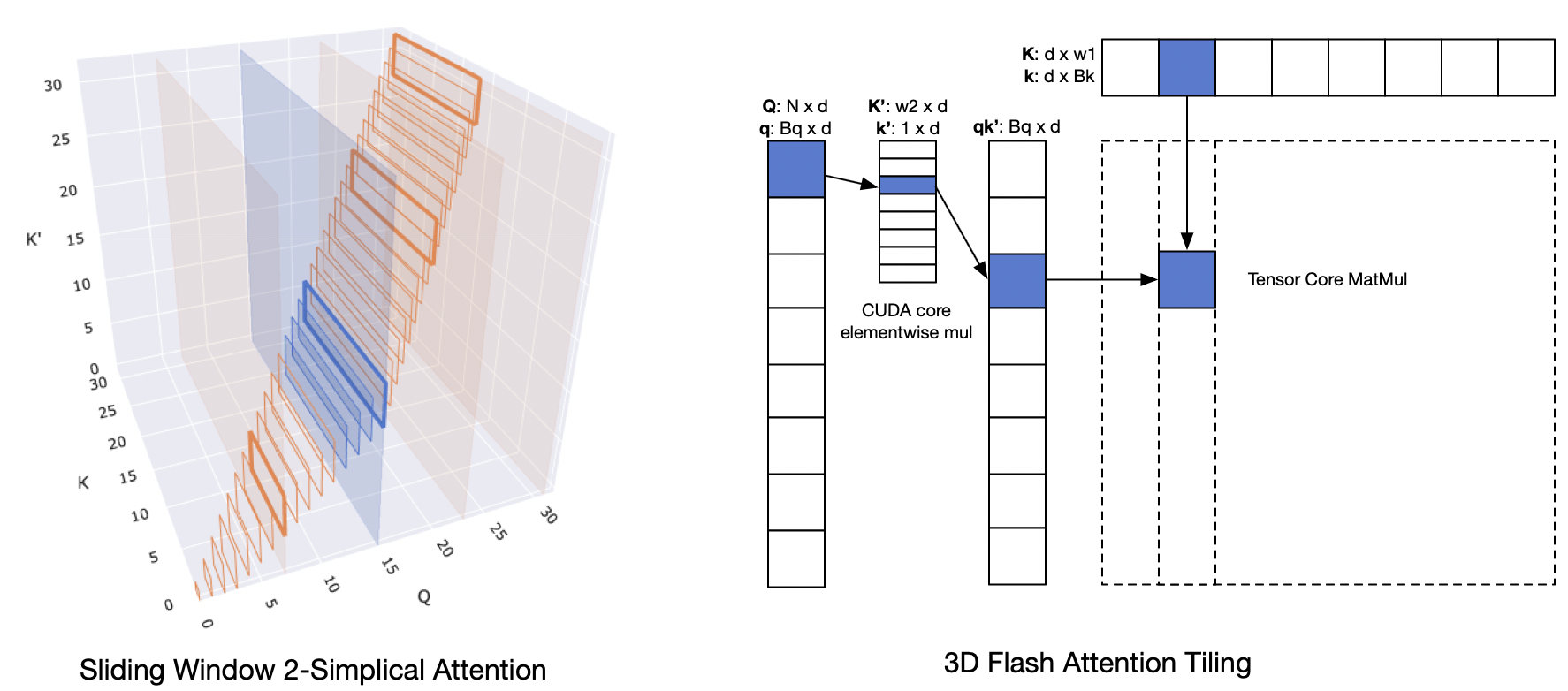}
\caption{\textbf{Left:} Visualization of sliding window 2-simplical attention. Each $Q_i$ attends to a $[w1, w2]$ shaped rectangle of $K$, $K'$. \textbf{Right:} Tiling to reduce 
2-simplicial einsum $QKK'$ to elementwise mul $QK'$ on CUDA core and tiled matmul $(QK')@K$ on tensor core.}
\label{fig:2-simplical-tiling}
\end{figure}

\begin{figure}[]
  \centering
  \begin{subfigure}[b]{0.45\textwidth}
    \centering
    \includegraphics[width=1.0\textwidth]{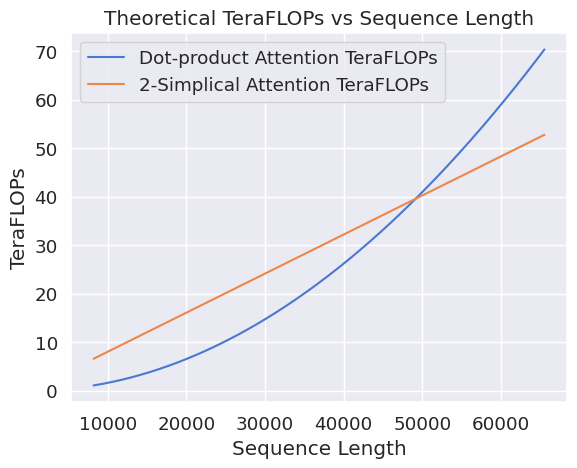}
  \end{subfigure}
  \begin{subfigure}[b]{0.45\textwidth}
      \centering
      \includegraphics[width=1.0\textwidth]{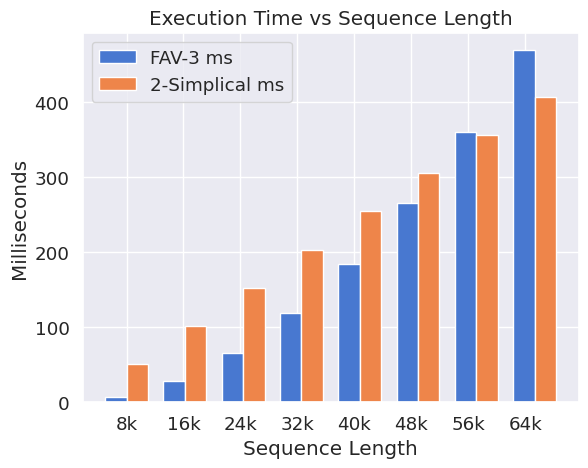}
  \end{subfigure}
  \caption{FLOPs and Latencies of FAv3 vs 2-simplical attention}
  \label{fig:fwd-latency}
\end{figure}

We introduce a series of kernel optimizatins tailored for 2-simplical attention, building off of Flash Attention \citep{dao2022flashattention} using online softmax. For the trilinear operations, we perform 2d tiling by merging one of the inputs via elementwise multiplication and executing matmul on the product as illustrated in Figure~\ref{fig:2-simplical-tiling}. This allows us to overlap both $QK$ and $VV'$ on CUDA Core with $(QK)@K'$ and $P @ (VV')$ on Tensor Core. Implementing this in Triton, we achieve 520 TFLOPS, rivaling the fastest FAv3 Triton implementations. Further optimization could be achieved with a lower-level language like CUTLASS for finer grained tuning and optimizations. Despite this, we achieve competitive performance compared to CUTLASS FAv3 for large sequence lengths, as shown in Figure~\ref{fig:fwd-latency}.

For the backwards pass, we have

\begin{equation}
    dV_{jd}=\sum_{i,k}\left(A_{ijk} \cdot dO_{id} \cdot V'_{kd}\right)
\end{equation}
\begin{equation}
    dV'_{kd}=\sum_{i,j}\left(A_{ijk}\cdot dO_{id} \cdot V_{jd}\right)
\end{equation}
\begin{equation}
    dP_{ijk}=\sum_d\left(dO_{id}\cdot V_{jd} \cdot V'_{kd}\right)
\end{equation}
\begin{equation}
    dS = dsoftmax_{jk}(dP)
\end{equation}
\begin{equation}
    dK_{jd} = \sum_{i,k}(Q_{id} \cdot dS_{ijk} \cdot K'_{kd})
\end{equation}
\begin{equation}
    dK'_{kd} = \sum_{i,k}(Q_{id} \cdot dS_{ijk} \cdot K_{jd})
\end{equation}
\begin{equation}
    dQ_{id} = \sum_{j,k}(dS_{ijk} \cdot K_{jd} \cdot K'_{kd})
\end{equation}

For the backwards pass, aggregations across three different dimension orderings introduces significant overhead from atomic operations. To mitigate this, we decompose the backward pass into two distinct kernels: one for computing $dK$ and $dV$, and another for $dK'$, $dV'$, and $dQ$. Although this approach incurs additional overhead from recomputing $O$ and $dS$, we find it is better than the extra overhead from atomics needed for a single fused kernel. We note this may be a limitation of Triton's coarser grained pipeline control making it difficult to hide the overhead from atomics.

For small $w_2$, we employ a two-stage approach to compute $dQ$ jointly with $dK'$, $dV'$ without atomics as detailed in Algorithm~\ref{alg:pseudo-code-bwd}. 
We divide $Q$ along the sequence dimension into $$[w_2, dim]$$ sized tiles. First we iterate over even tiles, storing $dQ$, $dK$, $dK'$, and $dV$, $dV'$. 
Then we iterate over odd tiles, storing $dQ$, and adding to $dK$, $dK'$ and $dV$, $dV'$.

\begin{algorithm}
  \caption{Backward pass for 2-simplicial attention}
  \label{alg:pseudo-code-bwd}
  \begin{algorithmic}[1]
    \Procedure{2-simplicial flash attention bwd}{$Q$, $K$, $V$, $K'$, $V'$, $w_1$, $w_2$}
    \For{\text{stage in [0, 1]}}
      \For{\text{q\_start in range(stage * $w_2$, seq\_len, $w_2$ * 2)}}
        \State $\mathrm{q\_end} \gets \mathrm{q\_start + w_2}$
        \For{\text{kv1\_start in range(q\_start - $w_1$, q\_end)}}
            \State $\mathrm{q\_tile} \gets \mathrm{Q[q\_start:q\_end]}$
            \State ...
            \State $\mathrm{k2\_tile} \gets\mathrm{K'[kv1\_start:q\_end]}$
            \State $dQ$ {+}= $dQ(\mathrm{q\_tile}, \mathrm{k2\_tile}, ...)$
            \State $dV'$ {+} = $dV'(\mathrm{q\_tile}, \mathrm{k2\_tile}, ...)$
            \State $dK'$ {+} = $dK'(\mathrm{q\_tile}, \mathrm{k2\_tile}, ...)$
          \EndFor
          \If{stage == 1}
            \State $dK'$ += \textbf{load} $dK'$
            \State $dV'$ += \textbf{load} $dV'$
          \EndIf
          \State \textbf{store} $dQ$, ..., $dK'$
      \EndFor
      \EndFor
    \EndProcedure
  \end{algorithmic}
\end{algorithm}

\section{Experiments \& Results}
We train a series of MoE models~\citep{jordan1994hierarchical,shazeer2017outrageously} 
ranging from $1$ billion active parameters and 
$57$ billion total parameters to $3.5$ billion active parameters and $176$ billion total parameters.
We use interleaved sliding-window 2-simplicial attention, where every fourth layer is a 2-simplicial attention layer.
The choice of this particular ordering is to distribute the load in attention
computation when using
pipeline parallelism \citep{huang2019gpipe,10.1145/3341301.3359646}, since
$2$-simplicial attention and global attention are the most compute intensive operations
in a single pipeline stage and have comparable FLOPs. 

We use the AdamW optimizer \citep{loshchilov2017fixing} with a peak
learning rate of $\num{4e-3}$ and weight decay of $0.0125$. We use a warmup of $4000$ steps
and use a cosine decay learning schedule decreasing the learning rate to $0.01\times$
of the peak learning rate. 
We report the negative log-likelihood on GSM8k~\citep{cobbe2021training},
MMLU~\citep{hendrycks2020measuring},
MMLU-pro~\citep{wang2024mmlu} and MBPP~\citep{austin2021program}, since these
benchmarks most strongly test math, reasoning and coding skills in pre-training.

\begin{table}[H]
\centering
\small
\begin{tabular}{lccccccc}
\toprule
\textbf{Model} & \textbf{Active Params} & \textbf{Total Params} & \textbf{GSM8k} & \textbf{MMLU} & \textbf{MMLU-pro} & \textbf{MBPP} \\
\midrule
Transformer   & ~1B & ~57B & 0.3277 & 0.6411 & 0.8718 & 0.2690 \\
2-simplicial  & ~1B & ~57B & 0.3302 & 0.6423 & 0.8718 & 0.2714 \\
$\Delta (\%)$ & & & \textcolor{red}{+0.79\%} & \textcolor{red}{+0.19\%} & \textcolor{ForestGreen}{-0.01\%} & \textcolor{red}{+0.88\%} \\
\hline
\addlinespace
Transformer                & 2B & ~100B & 0.2987 & 0.5932 & 0.8193 & 0.2435 \\
2-simplicial  & ~2B & ~100B & 0.2942 & 0.5862 & 0.8135 & 0.2411 \\
\addlinespace
$\Delta (\%)$ & & &\textcolor{ForestGreen}{-1.51\%} & \textcolor{ForestGreen}{-1.19\%} & \textcolor{ForestGreen}{-0.71\%} & \textcolor{ForestGreen}{-1\%} \\
\hline
\addlinespace
Transformer                & ~3.5B & ~176B & 0.2781 & 0.5543 & 0.7858 & 0.2203 \\
2-simplicial & ~3.5B & ~176B & 0.2718 & 0.5484 & 0.7689 & 0.2193 \\
\addlinespace
$\Delta (\%)$ & & &\textcolor{ForestGreen}{-2.27\%} & \textcolor{ForestGreen}{-1.06\%} & \textcolor{ForestGreen}{-2.15\%} & \textcolor{ForestGreen}{-0.45\%} \\
\bottomrule
\end{tabular}
\caption{
Negative log-likelihood of Transformer~\citep{vaswani2017attention} versus 2-simplicial attention.
For MMLU~\citep{hendrycks2020measuring} and MMLU-pro~\citep{wang2024mmlu} we measure the 
negative log-likelihood of the choice together with the entire answer. For 
GSM8k~\citep{cobbe2021training} we use $5$-shots for the results.
}
\label{tab:ablations}
\end{table}
We see that the decrease ($\Delta$) in negative log-likelihood scaling from a $1.0$ billion
(active) parameter model increases going to a $3.5$ billion (active) parameter model. Furthermore,
on models smaller than $2.0$ billion (active) parameters, we see no gains from using 
$2$-simplicial attention. From Table~\ref{tab:ablations} we can
estimate how the power law coefficients for the $2$-simplicial attention differ
from dot product attention. Recall from Section~\ref{sec:overview-scaling-laws} that the 
loss can be expressed as:
\begin{equation}
L(N, D) = E + \frac{A}{N^\alpha} + \frac{B}{D^\beta}.
\end{equation}
Since we train both the models on the same fixed number of tokens, we may ignore the third term
and simply write the loss as:
\begin{align}\label{eq:scaling-laws-param-only}
L(N) &= E' + \frac{A}{N^\alpha},\\
\log{L(N)} &\approx \log{E''} + \log{A} - \alpha \log{N}\\
-\log{L(N)} &= \alpha \log{N} + \beta,
\end{align}
where $\beta = - \log{E''} - log{A}$ and $E''$ is an approximation to $E'$ since $E'$ is small.
Note that here we used $\log({a + b}) = \log({1+a/b}) + \log(b)$ to separate out the two terms,
with the $1 + a/b$ term hidden in $E''$.
Therefore we can estimate $\alpha, \beta$ for both sets of models
from the losses in Table~\ref{tab:ablations} where we use for $N$ the active parameters in each model. 
We estimate the slope $\alpha$ and the intercept $\beta$ for both the Transformer as well as the 
$2$-simplicial Transformer in Table~\ref{tab:power-law}. We see that
$2$-simplicial attention has a steeper slope $\alpha$, i.e. a higher exponent in its scaling law
compared to dot product attention Transformer~\citep{vaswani2017attention}.

\begin{table}[H]
  \centering
  \small
  \begin{tabular}{lcccccccc}
    \toprule
    \textbf{Model}  & \multicolumn{2}{c}{\textbf{GSM8k}} & \multicolumn{2}{c}{\textbf{MMLU}} & \multicolumn{2}{c}{\textbf{MMLU-pro}} & \multicolumn{2}{c}{\textbf{MBPP}} \\
    & $\alpha$  & $\beta$ & $\alpha$ & $\beta$ & $\alpha$ & $\beta$ & $\alpha$ & $\beta$ \\
    \midrule
    Transformer   & 0.1420 & -1.8280 & 0.1256 & -2.1606 & 0.0901 & -1.7289 & 0.1720 & -2.2569 \\
    2-simplicial   & 0.1683 & -2.3939 & 0.1364 & -2.3960 & 0.1083 & -2.1181 & 0.1837 & -2.5201 \\
    $\Delta (\%)$ & \textcolor{ForestGreen}{18.5\%} & &\textcolor{ForestGreen}{8.5\%} & & \textcolor{ForestGreen}{20.2\%} & & \textcolor{ForestGreen}{6.8\%} & \\
    \bottomrule
  \end{tabular}
\caption{
Estimates of the power law coefficients $\alpha$ and $\beta$ for the Transformer~\citep{vaswani2017attention} and 
2-simplicial attention.
}
\label{tab:power-law}
\end{table}

\begin{table}[H]
  \centering
  \small
  \begin{tabular}{lcccccccc}
    \toprule
    \textbf{Model}  & \multicolumn{2}{c}{\textbf{GSM8k}} & \multicolumn{2}{c}{\textbf{MMLU}} & \multicolumn{2}{c}{\textbf{MMLU-pro}} & \multicolumn{2}{c}{\textbf{MBPP}} \\
    & $R^2$  & residual & $R^2$ & residual & $R^2$ & residual & $R^2$ & residual \\
    \midrule
    Transformer   & 0.9998 & $\num{2.8e-06}$ & 0.9995 & \num{4.7e-06} & 0.9972 & \num{1.5e-05} & 0.9962 & $\num{7.5e-05}$ \\
    2-simplicial   & 0.9974 & \num{4.9e-05} & 0.9989 & \num{1.3e-05} & 0.9999 & \num{4.6e-08} & 0.9999 & \num{1.5e-06} \\
    \bottomrule
  \end{tabular}
\caption{$R^2$ and residuals measuring goodness of fit for Table~\ref{tab:power-law}.}
\end{table}


\section{Discussion}
While $2$-simplicial attention improves the exponent
in the scaling laws, we should caveat that 
the technique maybe more useful when we are in the regime when
token efficiency becomes more important.
Our Triton kernel while efficient for prototyping is still
far away from being used in production. More work in 
co-designing the implementation of $2$-simplicial attention
tailored to the specific hardware accelerator
is needed in the future.

\section{Conclusion}
We show that a similar sized $2$-simplicial attention~\citep{clift2019logic} 
improves on dot product 
attention of \cite{vaswani2017attention} by improving the negative log likelihood
on reasoning, math and coding problems (see Table~\ref{tab:ablations}). 
We quantify this
explicitly in Table~\ref{tab:power-law} by demonstrating that $2$-simplicial attention
changes the exponent corresponding to parameters in the scaling law of
Equation~\ref{eq:scaling-laws-param-only}: in 
particular it has a higher $\alpha$ for reasoning and coding tasks compared to the 
Transformer~\citep{vaswani2017attention} which leads to more favorable scaling under
token constraints. Furthermore, the percentage increase in the scaling exponent $\alpha$ 
is higher for less saturated and more challenging benchmarks such as MMLU-pro and GSM8k.

We hope that scaling 
$2$-simplicial Transformers could unlock significant improvements in downstream 
performance on reasoning-heavy tasks, helping to overcome the current limitations of pre-
training scalability. Furthermore, we believe that developing a specialized and efficient implementation is key to fully unlocking the potential of this architecture.

\section{Acknowledgments}
The authors gratefully acknowledge the invaluable support and feedback from 
Chuanhao Zhuge, Tony Liu, Ying Zhang, Ajit Mathews, Afroz Mohiuddin, Vinay Rao and Dhruv 
Choudhary.

\bibliography{deeplearn}
\bibliographystyle{iclr2025_conference}

\appendix
\newpage
\section{Rotation invariant trilinear forms}\label{sec:proof-rotation-invariance}
\subsection{Proof for Theorem \ref{thm:det}}
We define the embedding functions for the Query and Key vectors such that their interaction within the Sum-of-Determinants attention mechanism computes the $\texttt{Match3}$ function. To handle cases where no match exists, we use a 7-dimensional embedding where the 7th dimension acts as a selector for a "blank pair" option, a technique adapted from  \texttt{Match2} construction in \cite{sanford2023representational}.

The construction for regular token pairs is based on the mathematical identity:
\begin{align}
\label{eq:costheta}
   \cos(\theta_1 + \theta_2 + \theta_3) = \det(M_1) + \det(-M_2),
\end{align}
where the matrices $M_1, M_2\in \mathbb{R}^{3\times 3}$ are defined as:
$$ M_1 = \begin{pmatrix} \cos(\theta_1) & \sin(\theta_1) & 0 \\ \sin(\theta_2) & \cos(\theta_2) & 0 \\ 0 & 0 & \cos(\theta_3) \end{pmatrix}, \quad -M_2 = \begin{pmatrix} -\sin(\theta_1) & \cos(\theta_1) & 0 \\ -\sin(\theta_2) & -\cos(\theta_2) & 0 \\ 0 & 0 & -\sin(\theta_3) \end{pmatrix} $$
Let $\theta_k = \frac{2\pi x_k}{M}$. We define the 7-dimensional query vector $\vq_i$ and key vectors $\vk_{j_1}, \vk'_{j_2}$ via an input MLP $\phi$ and matrices $Q, K, K'$. Let $c$ be a large scaling constant.

The 7-dimensional query vector $q_i = Q\phi(x_i)$ is defined as:
$$ \vq_i = (c\cos(\theta_i), c\sin(\theta_i), 0, -c\sin(\theta_i), c\cos(\theta_i), 0, c) $$
The key vectors $\vk_{j_1} = K\phi(x_{j_1})$ and $\vk'_{j_2} = K'\phi(x_{j_2})$ for regular tokens are 
defined as:
$$ \vk_{j_1} = (\sin(\theta_{j_1}), \cos(\theta_{j_1}), 0, -\sin(\theta_{j_1}), -\cos(\theta_{j_1}), 0, 0) $$
$$ \vk'_{j_2} = (0, 0, \cos(\theta_{j_2}), 0, 0, -\sin(\theta_{j_2}), 0) $$
The attention score is computed via a hybrid mechanism:
\begin{enumerate}
    \item \textbf{For regular pairs $(j_1, j_2)$}, the score is the sum of determinants of two 3D chunks formed from the first 6 dimensions of the vectors. The 7th dimension of the keys is 0, so it is ignored in this term.
    \begin{align*}
    A_{i, j_1, j_2} &= \det(\vq_i[:3], k_{j_1}[:3], \vk_{j_2}'[:3]) + \det(\vq_i[3:6], k_{j_1}[3:6], \vk_{j_2}'[3:6])\\
    &= c\cdot (\det(M_1) + \det(-M_2))\quad (\text{from }(\ref{eq:costheta}))\\
    &=c \cdot \cos\left(\frac{2\pi(x_i + x_{j_1} + x_{j_2})}{M}\right)\quad (\text{since }\theta_i = 2\pi x_k/M ),
    \end{align*}
    where $\vq_i[l:l+m] = \{(\vq_i)_l,\ldots,(\vq_i)_{l+m-1}\}$, denotes array slicing. 
    \item \textbf{For the blank pair}, the score is computed using the 7th dimension. It is the dot product of the query vector $\vq_i$ and a fixed key vector $\vk_{\text{blank}} = (0,0,0,0,0,0,1)$:
    $$A_{i,\text{blank}} = \vq_i \cdot \vk_{\text{blank}} = c$$
\end{enumerate}

As a result, the attention score is maximized to a value of $c$ if and only if $x_i + x_{j_1} + x_{j_2} = 0 \pmod M$. The blank pair also receives a score of $c$. For any non-matching triple, the score is strictly less than $c$.

The value vectors are defined by matrices $V$ and $V'$.
\begin{itemize}
    \item For any \textbf{regular token} $x_j$, we set its value embeddings to be $V\phi(x_j) = 1$ and $V'\phi(x_j) = 1$. The resulting value for the pair $(j_1, j_2)$ in the final value matrix is their Kronecker product, which is 1.
    \item For the \textbf{blank pair}, the corresponding value is 0.
\end{itemize}

Let $\beta_i$ be the number of pairs $(j_1, j_2)$ that form a match with $x_i$. The softmax function distributes the attention weight almost exclusively among the entries with a score of $c$.
\begin{itemize}
    \item If no match exists ($\beta_i=0$), the blank pair receives all the attention, and the output is $\approx 0$ since its value is 0.
    \item If at least one match exists ($\beta_i \ge 1$), the attention is distributed among the $\beta_i$ matching pairs and the 1 blank pair. The output of the attention layer will be approximately $\frac{\beta_i \cdot (1) + 1 \cdot (0)}{\beta_i + 1} = \frac{\beta_i}{\beta_i+1}$.
\end{itemize}

The final step is to design an output MLP $\psi$ such that $\psi(z)=1$ if $z \ge 1/2$ and $\psi(z)=0$ otherwise, which is straightforward to implement.

\section{Triton Kernel: Forward pass for 2-simplicial Attention}
\begin{lstlisting}[language=Python, caption={Forward pass for 2-simplicial attention.}]
@triton.autotune(
    configs=[
        Config(
            {
                "BLOCK_SIZE_Q": 64,
                "BLOCK_SIZE_KV": 32,
                "num_stages": 1,
            },
            num_warps=4,
        )
    ],
    key=["HEAD_DIM"],
)
@triton.jit
def two_simplicial_attn_fwd_kernel(
    Q_ptr,  # [b, s, k, h]
    K1_ptr,  # [b, s, k, h]
    K2_ptr,  # [b, s, k, h]
    V1_ptr,  # [b, s, k, h]
    V2_ptr,  # [b, s, k, h]
    O_ptr,  # [b, s, k, h]
    M_ptr,  # [b, k, s]
    bs,
    seq_len,
    num_heads,
    head_dim,
    w1: tl.constexpr,
    w2: tl.constexpr,
    q_stride_b,
    q_stride_s,
    q_stride_k,
    q_stride_h,
    k1_stride_b,
    k1_stride_s,
    k1_stride_k,
    k1_stride_h,
    k2_stride_b,
    k2_stride_s,
    k2_stride_k,
    k2_stride_h,
    v1_stride_b,
    v1_stride_s,
    v1_stride_k,
    v1_stride_h,
    v2_stride_b,
    v2_stride_s,
    v2_stride_k,
    v2_stride_h,
    out_stride_b,
    out_stride_s,
    out_stride_k,
    out_stride_h,
    m_stride_b,
    m_stride_k,
    m_stride_s,
    BLOCK_SIZE_Q: tl.constexpr,
    BLOCK_SIZE_KV: tl.constexpr,
    HEAD_DIM: tl.constexpr,
    INPUT_PRECISION: tl.constexpr,
    SM_SCALE: tl.constexpr,
    K2_BIAS: tl.constexpr,
    V2_BIAS: tl.constexpr,
    num_stages: tl.constexpr,
):
    data_dtype = tl.bfloat16
    compute_dtype = tl.float32
    gemm_dtype = tl.bfloat16

    q_start = tl.program_id(0) * BLOCK_SIZE_Q
    q_end = q_start + BLOCK_SIZE_Q
    bk = tl.program_id(1)
    offs_b = bk // num_heads
    offs_k = bk % num_heads

    qkv_offs_bk = offs_b * q_stride_b + offs_k * q_stride_k

    Q_ptr += qkv_offs_bk
    K1_ptr += qkv_offs_bk
    K2_ptr += qkv_offs_bk
    V1_ptr += qkv_offs_bk
    V2_ptr += qkv_offs_bk
    O_ptr += qkv_offs_bk
    M_ptr += offs_b * m_stride_b + offs_k * m_stride_k

    m_i = tl.zeros((BLOCK_SIZE_Q,), dtype=compute_dtype) - float("inf")
    l_i = tl.zeros((BLOCK_SIZE_Q,), dtype=compute_dtype)
    acc = tl.zeros((BLOCK_SIZE_Q, HEAD_DIM), dtype=compute_dtype)

    q_offs_s = q_start + tl.arange(0, BLOCK_SIZE_Q)
    qkv_offs_h = tl.arange(0, HEAD_DIM)
    q_mask_s = q_offs_s < seq_len
    qkv_mask_h = qkv_offs_h < head_dim
    q_offs = q_offs_s[:, None] * q_stride_s + qkv_offs_h[None, :] * q_stride_h
    q_mask = q_mask_s[:, None] & (qkv_mask_h[None, :])

    q_tile = tl.load(Q_ptr + q_offs, mask=q_mask).to(
        compute_dtype
    )  # [BLOCK_SIZE_Q, HEAD_DIM]
    softmax_scale = tl.cast(SM_SCALE, gemm_dtype)

    for kv1_idx in tl.range(tl.maximum(0, q_start - w1), tl.minimum(seq_len, q_end)):
        k1_offs = kv1_idx * k1_stride_s + qkv_offs_h * k1_stride_h
        k1_tile = (tl.load(K1_ptr + k1_offs, mask=qkv_mask_h).to(compute_dtype))[
            None, :
        ]  # [1, HEAD_DIM]
        qk1 = q_tile * k1_tile  # [BLOCK_SIZE_Q, HEAD_DIM]
        qk1 = qk1.to(gemm_dtype)

        v1_offs = kv1_idx * v1_stride_s + qkv_offs_h * v1_stride_h
        v1_tile = (tl.load(V1_ptr + v1_offs, mask=qkv_mask_h).to(compute_dtype))[
            None, :
        ]  # [1, HEAD_DIM]

        for kv2_idx in tl.range(
            tl.maximum(0, q_start - w2),
            tl.minimum(seq_len, q_end),
            BLOCK_SIZE_KV,
            num_stages=num_stages,
        ):
            kv2_offs_s = kv2_idx + tl.arange(0, BLOCK_SIZE_KV)
            kv2_mask_s = kv2_offs_s < seq_len
            k2t_mask = kv2_mask_s[None, :] & qkv_mask_h[:, None]
            v2_mask = kv2_mask_s[:, None] & qkv_mask_h[None, :]
            k2_offs = (
                kv2_offs_s[None, :] * k2_stride_s + qkv_offs_h[:, None] * k2_stride_h
            )
            v2_offs = (
                kv2_offs_s[:, None] * v2_stride_s + qkv_offs_h[None, :] * v2_stride_h
            )
            k2t_tile = tl.load(K2_ptr + k2_offs, mask=k2t_mask).to(
                compute_dtype
            )  # [HEAD_DIM, BLOCK_SIZE_KV]
            v2_tile = tl.load(V2_ptr + v2_offs, mask=v2_mask).to(
                compute_dtype
            )  # [BLOCK_SIZE_KV, HEAD_DIM]
            k2t_tile += K2_BIAS
            v2_tile += V2_BIAS
            k2t_tile = k2t_tile.to(gemm_dtype)
            v2_tile = v2_tile.to(compute_dtype)

            qk = tl.dot(
                qk1 * softmax_scale,
                k2t_tile,
                input_precision="tf32",  # INPUT_PRECISION,
                out_dtype=tl.float32,
            )  # [BLOCK_SIZE_Q, BLOCK_SIZE_KV]

            qk_mask = q_mask_s[:, None] & kv2_mask_s[None, :]
            # Mask for q_idx - w1 < kv1_idx <= q_idx
            # and q_idx - w2 < kv2_offs_s <= q_idx
            kv1_local_mask = ((q_offs_s[:, None] - w1) < kv1_idx) & (
                kv1_idx <= q_offs_s[:, None]
            )
            kv2_local_mask = ((q_offs_s[:, None] - w2) < kv2_offs_s[None, :]) & (
                kv2_offs_s[None, :] <= q_offs_s[:, None]
            )
            qk_mask &= kv1_local_mask & kv2_local_mask
            qk += tl.where(qk_mask, 0, -1.0e38)

            m_ij = tl.maximum(m_i, tl.max(qk, 1))
            p = tl.math.exp(qk - m_ij[:, None])
            l_ij = tl.sum(p, 1)
            alpha = tl.math.exp(m_i - m_ij)
            l_i = l_i * alpha + l_ij
            acc = acc * alpha[:, None]

            v12_tile = v1_tile * v2_tile  # [BLOCK_SIZE_KV, HEAD_DIM]
            acc += tl.dot(
                p.to(gemm_dtype),
                v12_tile.to(gemm_dtype),
                input_precision="ieee",  # INPUT_PRECISION,
                out_dtype=tl.float32,
            )

            m_i = m_ij
    acc = acc / l_i[:, None]

    acc = tl.where(q_mask, acc, 0.0)
    acc = acc.to(data_dtype)
    out_offs = q_offs_s[:, None] * out_stride_s + qkv_offs_h[None, :] * out_stride_h
    tl.store(O_ptr + out_offs, acc, mask=q_mask)

    m = m_i + tl.log(l_i)

    m_offs = q_offs_s * m_stride_s
    m_mask = q_offs_s < seq_len
    tl.store(M_ptr + m_offs, m, mask=m_mask)
\end{lstlisting}

\section{Triton Kernel: Backward pass for 2-simplicial Attention}
\begin{lstlisting}[language=Python, caption={Backward pass for 2-simplicial attention.}]
@triton.jit
def two_simplicial_attn_bwd_kv1_kernel(
    Q_ptr,  # [b, s, k, h]
    K1_ptr,  # [b, s, k, h]
    K2_ptr,  # [b, s, k, h]
    V1_ptr,  # [b, s, k, h]
    V2_ptr,  # [b, s, k, h]
    dO_ptr,  # [b, s, k, h]
    M_ptr,  # [b, k, s]
    D_ptr,  # [b, k, s]
    dQ_ptr,  # [b, s, k, h]
    dK1_ptr,  # [b, s, k, h]
    dV1_ptr,  # [b, s, k, h]
    # Skip writing dk2, dv2 for now.
    bs,
    seq_len,
    num_heads,
    head_dim,
    w1,  # Q[i]: KV1(i-w1,i]
    w2,  # Q[i]: KV2(i-w2,i]
    q_stride_b,
    q_stride_s,
    q_stride_k,
    q_stride_h,
    k1_stride_b,
    k1_stride_s,
    k1_stride_k,
    k1_stride_h,
    k2_stride_b,
    k2_stride_s,
    k2_stride_k,
    k2_stride_h,
    v1_stride_b,
    v1_stride_s,
    v1_stride_k,
    v1_stride_h,
    v2_stride_b,
    v2_stride_s,
    v2_stride_k,
    v2_stride_h,
    dO_stride_b,
    dO_stride_s,
    dO_stride_k,
    dO_stride_h,
    m_stride_b,
    m_stride_k,
    m_stride_s,
    d_stride_b,
    d_stride_k,
    d_stride_s,
    dq_stride_b,
    dq_stride_s,
    dq_stride_k,
    dq_stride_h,
    dk1_stride_b,
    dk1_stride_s,
    dk1_stride_k,
    dk1_stride_h,
    dv1_stride_b,
    dv1_stride_s,
    dv1_stride_k,
    dv1_stride_h,
    BLOCK_SIZE_Q: tl.constexpr,
    BLOCK_SIZE_KV: tl.constexpr,
    HEAD_DIM: tl.constexpr,
    SM_SCALE: tl.constexpr,
    K2_BIAS: tl.constexpr,
    V2_BIAS: tl.constexpr,
    COMPUTE_DQ: tl.constexpr,
    num_stages: tl.constexpr,
    is_flipped: tl.constexpr,
):
    data_dtype = tl.bfloat16
    compute_dtype = tl.float32
    gemm_dtype = tl.bfloat16

    kv1_start = tl.program_id(0) * BLOCK_SIZE_KV
    kv1_end = kv1_start + BLOCK_SIZE_KV
    bk = tl.program_id(1)
    offs_b = bk // num_heads
    offs_k = bk % num_heads

    qkv_offs_bk = offs_b * q_stride_b + offs_k * q_stride_k
    Q_ptr += qkv_offs_bk
    K1_ptr += qkv_offs_bk
    K2_ptr += qkv_offs_bk
    V1_ptr += qkv_offs_bk
    V2_ptr += qkv_offs_bk

    dO_ptr += offs_b * dO_stride_b + offs_k * dO_stride_k
    M_ptr += offs_b * m_stride_b + offs_k * m_stride_k
    D_ptr += offs_b * d_stride_b + offs_k * d_stride_k
    dK1_ptr += offs_b * dk1_stride_b + offs_k * dk1_stride_k
    dV1_ptr += offs_b * dv1_stride_b + offs_k * dv1_stride_k
    if COMPUTE_DQ:
        dQ_ptr += offs_b * dq_stride_b + offs_k * dq_stride_k

    softmax_scale = tl.cast(SM_SCALE, gemm_dtype)
    qkv_offs_h = tl.arange(0, HEAD_DIM)
    qkv_mask_h = qkv_offs_h < head_dim

    kv1_offs_s = kv1_start + tl.arange(0, BLOCK_SIZE_KV)

    k1_offs = kv1_offs_s[:, None] * k1_stride_s + qkv_offs_h[None, :] * k1_stride_h
    kv1_mask_s = kv1_offs_s < seq_len
    kv1_mask = kv1_mask_s[:, None] & qkv_mask_h[None, :]
    k1_tile = tl.load(K1_ptr + k1_offs, mask=kv1_mask).to(
        compute_dtype
    )  # [BLOCK_SIZE_KV, HEAD_DIM]
    v1_offs = kv1_offs_s[:, None] * v1_stride_s + qkv_offs_h[None, :] * v1_stride_h
    v1_tile = tl.load(V1_ptr + v1_offs, mask=kv1_mask).to(
        compute_dtype
    )  # [BLOCK_SIZE_KV, HEAD_DIM]
    if is_flipped:
        k1_tile += K2_BIAS
        v1_tile += V2_BIAS
    dv1 = tl.zeros((BLOCK_SIZE_KV, HEAD_DIM), compute_dtype)
    dk1 = tl.zeros((BLOCK_SIZE_KV, HEAD_DIM), compute_dtype)
    # for kv2_idx in tl.range(0, seq_len):
    # kv1 - w2 < kv2 <= kv1 + w1
    for kv2_idx in tl.range(
        tl.maximum(0, kv1_start - w2), tl.minimum(seq_len, kv1_end + w1)
    ):
        k2_offs = kv2_idx * k2_stride_s + qkv_offs_h * k2_stride_h
        k2_tile = (tl.load(K2_ptr + k2_offs, mask=qkv_mask_h).to(compute_dtype))[
            None, :
        ]  # [1, HEAD_DIM]
        v2_offs = kv2_idx * v2_stride_s + qkv_offs_h * v2_stride_h
        v2_tile = (tl.load(V2_ptr + v2_offs, mask=qkv_mask_h).to(compute_dtype))[
            None, :
        ]  # [1, HEAD_DIM]
        if not is_flipped:
            k2_tile += K2_BIAS
            v2_tile += V2_BIAS
        k1k2 = k1_tile * k2_tile  # [BLOCK_SIZE_KV, HEAD_DIM]
        v1v2 = v1_tile * v2_tile  # [BLOCK_SIZE_KV, HEAD_DIM]
        k1k2 = k1k2.to(gemm_dtype)
        v1v2 = v1v2.to(gemm_dtype)
        # kv1 <= q < kv1 + w1
        # kv2 <= q < kv2 + w2
        q_start = tl.maximum(kv1_start, kv2_idx)
        q_end = tl.minimum(seq_len, tl.minimum(kv1_end + w1, kv2_idx + w2))
        for q_idx in tl.range(q_start, q_end, BLOCK_SIZE_Q):
            # Load qt, m, d, dO
            q_offs_s = q_idx + tl.arange(0, BLOCK_SIZE_Q)
            q_offs = q_offs_s[None, :] * q_stride_s + qkv_offs_h[:, None] * q_stride_h
            q_mask_s = q_offs_s < seq_len
            qt_mask = q_mask_s[None, :] & qkv_mask_h[:, None]
            qt_tile = tl.load(Q_ptr + q_offs, mask=qt_mask).to(
                gemm_dtype
            )  # [HEAD_DIM, BLOCK_SIZE_Q]
            m_offs = q_offs_s * m_stride_s
            m_tile = tl.load(M_ptr + m_offs, mask=q_mask_s).to(compute_dtype)[
                None, :
            ]  # [1, BLOCK_SIZE_Q]
            d_offs = q_offs_s * d_stride_s
            d_tile = tl.load(D_ptr + d_offs, mask=q_mask_s).to(compute_dtype)[
                None, :
            ]  # [1, BLOCK_SIZE_Q]
            dO_offs = (
                q_offs_s[:, None] * dO_stride_s + qkv_offs_h[None, :] * dO_stride_h
            )
            dO_tile = tl.load(
                dO_ptr + dO_offs, mask=q_mask_s[:, None] & qkv_mask_h[None, :]
            ).to(compute_dtype)  # [BLOCK_SIZE_Q, HEAD_DIM]
            if COMPUTE_DQ:
                dq = tl.zeros((BLOCK_SIZE_Q, HEAD_DIM), tl.float32)
            # Compute dv1.
            # [KV, D] @ [D, Q] => [KV, Q]
            qkkT = tl.dot(
                k1k2, qt_tile * softmax_scale, out_dtype=tl.float32
            )  # [BLOCK_SIZE_KV, BLOCK_SIZE_Q]

            # Mask qkkT to -inf.
            kv1_local_mask = ((q_offs_s[None, :] - w1) < kv1_offs_s[:, None]) & (
                kv1_offs_s[:, None] <= q_offs_s[None, :]
            )
            kv2_local_mask = ((q_offs_s - w2) < kv2_idx) & (kv2_idx <= q_offs_s)
            local_mask = (
                kv1_local_mask & kv2_local_mask[None, :]
            )  # [BLOCK_SIZE_KV, BLOCK_SIZE_Q]
            qkkT = tl.where(local_mask, qkkT, -1.0e38)

            pT = tl.exp(qkkT - m_tile)  # [BLOCK_SIZE_KV, BLOCK_SIZE_Q]
            pT = tl.where(local_mask, pT, 0.0)
            dOv2 = dO_tile * v2_tile  # [BLOCK_SIZE_Q, HEAD_DIM]
            dv1 += tl.dot(
                pT.to(gemm_dtype), dOv2.to(gemm_dtype), out_dtype=tl.float32
            )  # [BLOCK_SIZE_KV, HEAD_DIM]
            
            dpT = tl.dot(
                v1v2, tl.trans(dO_tile.to(gemm_dtype)), out_dtype=tl.float32
            )  # [BLOCK_SIZE_KV, BLOCK_SIZE_Q]
            dsT = pT * (dpT - d_tile)  # [BLOCK_SIZE_KV, BLOCK_SIZE_Q]
            dsT = tl.where(local_mask, dsT, 0.0)
            dsT = dsT.to(gemm_dtype)

            dk1 += (
                tl.dot(dsT, tl.trans(qt_tile), out_dtype=tl.float32)
                * k2_tile.to(tl.float32)
                * softmax_scale
            )
            if COMPUTE_DQ:
                # dq[q, d] = dsT.T[q, kv1] @ k1k2[kv1, d]
                dq += (
                    tl.dot(tl.trans(dsT), k1k2, out_dtype=tl.float32) * softmax_scale
                )  # [BLOCK_SIZE_Q, HEAD_DIM]
                dq_offs = (
                    q_offs_s[:, None] * dq_stride_s + qkv_offs_h[None, :] * dq_stride_h
                )   
                tl.atomic_add(
                    dQ_ptr + dq_offs, dq, mask=q_mask_s[:, None] & qkv_mask_h[None, :]
                )
    dv1_offs = kv1_offs_s[:, None] * dv1_stride_s + qkv_offs_h[None, :] * dv1_stride_h
    dk1_offs = kv1_offs_s[:, None] * dk1_stride_s + qkv_offs_h[None, :] * dk1_stride_h
    tl.store(dV1_ptr + dv1_offs, dv1.to(data_dtype), mask=kv1_mask)
    tl.store(dK1_ptr + dk1_offs, dk1.to(data_dtype), mask=kv1_mask)
\end{lstlisting}

\begin{lstlisting}[language=Python, caption={Backward pass for 2-simplicial 
attention optimized for small $w_2$ avoiding atomic adds.}]
@triton.autotune(
    configs=[
        Config(
            {
                "BLOCK_SIZE_Q": 32,
                "BLOCK_SIZE_KV2": 64,
                "num_stages": 1,
            },
            num_warps=4,
        )
    ],
    key=["HEAD_DIM"],
)
@triton.jit
def two_simplicial_attn_bwd_kv2q_kernel(
    Q_ptr,  # [b, s, k, h]
    K1_ptr,  # [b, s, k, h]
    K2_ptr,  # [b, s, k, h]
    V1_ptr,  # [b, s, k, h]
    V2_ptr,  # [b, s, k, h]
    dO_ptr,  # [b, s, k, h]
    M_ptr,  # [b, k, s]
    D_ptr,  # [b, k, s]
    dQ_ptr,  # [b, s, k, h]
    dK2_ptr,  # [b, s, k, h]
    dV2_ptr,  # [b, s, k, h]
    bs,
    seq_len,
    num_heads,
    head_dim,
    w1,  # Q[i]: KV1(i-w1,i]
    w2,  # Q[i]: KV2(i-w2,i]
    q_stride_b,
    q_stride_s,
    q_stride_k,
    q_stride_h,
    k1_stride_b,
    k1_stride_s,
    k1_stride_k,
    k1_stride_h,
    k2_stride_b,
    k2_stride_s,
    k2_stride_k,
    k2_stride_h,
    v1_stride_b,
    v1_stride_s,
    v1_stride_k,
    v1_stride_h,
    v2_stride_b,
    v2_stride_s,
    v2_stride_k,
    v2_stride_h,
    dO_stride_b,
    dO_stride_s,
    dO_stride_k,
    dO_stride_h,
    m_stride_b,
    m_stride_k,
    m_stride_s,
    d_stride_b,
    d_stride_k,
    d_stride_s,
    dq_stride_b,
    dq_stride_s,
    dq_stride_k,
    dq_stride_h,
    dk2_stride_b,
    dk2_stride_s,
    dk2_stride_k,
    dk2_stride_h,
    dv2_stride_b,
    dv2_stride_s,
    dv2_stride_k,
    dv2_stride_h,
    BLOCK_SIZE_Q: tl.constexpr,
    BLOCK_SIZE_KV2: tl.constexpr,
    HEAD_DIM: tl.constexpr,
    SM_SCALE: tl.constexpr,
    K2_BIAS: tl.constexpr,
    V2_BIAS: tl.constexpr,
    num_stages: tl.constexpr,
    IS_SECOND_PASS: tl.constexpr,
):
    assert BLOCK_SIZE_KV2 == BLOCK_SIZE_Q + w2
    data_dtype = tl.bfloat16
    compute_dtype = tl.float32
    gemm_dtype = tl.bfloat16

    # First pass does even tiles, second pass does odd tiles.
    q_start = tl.program_id(0) * BLOCK_SIZE_KV2
    if IS_SECOND_PASS:
        q_start += BLOCK_SIZE_Q
    q_end = q_start + BLOCK_SIZE_Q
    kv2_start = q_start - w2

    bk = tl.program_id(1)
    offs_b = bk // num_heads
    offs_k = bk % num_heads

    qkv_offs_bk = offs_b * q_stride_b + offs_k * q_stride_k
    Q_ptr += qkv_offs_bk
    K1_ptr += qkv_offs_bk
    K2_ptr += qkv_offs_bk
    V1_ptr += qkv_offs_bk
    V2_ptr += qkv_offs_bk
    
    dO_ptr += offs_b * dO_stride_b + offs_k * dO_stride_k
    M_ptr += offs_b * m_stride_b + offs_k * m_stride_k
    D_ptr += offs_b * d_stride_b + offs_k * d_stride_k
    dQ_ptr += offs_b * dq_stride_b + offs_k * dq_stride_k
    dK2_ptr += offs_b * dk2_stride_b + offs_k * dk2_stride_k
    dV2_ptr += offs_b * dv2_stride_b + offs_k * dv2_stride_k

    softmax_scale = tl.cast(SM_SCALE, gemm_dtype)
    qkv_offs_h = tl.arange(0, HEAD_DIM)
    qkv_mask_h = qkv_offs_h < head_dim

    q_offs_s = q_start + tl.arange(0, BLOCK_SIZE_Q)
    kv2_offs_s = kv2_start + tl.arange(0, BLOCK_SIZE_KV2)
    q_offs = q_offs_s[:, None] * q_stride_s + qkv_offs_h[None, :] * q_stride_h
    kv2_offs = kv2_offs_s[:, None] * k2_stride_s + qkv_offs_h[None, :] * k2_stride_h
    m_offs = q_offs_s * m_stride_s
    d_offs = q_offs_s * d_stride_s
    dO_offs = q_offs_s[:, None] * dO_stride_s + qkv_offs_h[None, :] * dO_stride_h
    q_mask_s = q_offs_s < seq_len
    q_mask = q_mask_s[:, None] & qkv_mask_h[None, :]
    kv2_mask_s = 0 <= kv2_offs_s and kv2_offs_s < seq_len
    kv2_mask = kv2_mask_s[:, None] & qkv_mask_h[None, :]

    q_tile = tl.load(Q_ptr + q_offs, mask=q_mask).to(
        compute_dtype
    )  # [BLOCK_SIZE_Q, HEAD_DIM]
    k2_tile = tl.load(K2_ptr + kv2_offs, mask=kv2_mask).to(gemm_dtype) # [KV2, HEAD_DIM]
    v2_tile = tl.load(V2_ptr + kv2_offs, mask=kv2_mask).to(gemm_dtype) # [KV2, HEAD_DIM]
    m_tile = tl.load(M_ptr + m_offs, mask=q_mask_s).to(compute_dtype) # [BLOCK_SIZE_Q]
    d_tile = tl.load(D_ptr + d_offs, mask=q_mask_s).to(compute_dtype) # [BLOCK_SIZE_Q]
    dO_tile = tl.load(dO_ptr + dO_offs, mask=q_mask).to(
        gemm_dtype
    )  # [BLOCK_SIZE_Q, HEAD_DIM]

    # Apply KV2 norm.
    k2_tile += K2_BIAS
    v2_tile += V2_BIAS
    k2_tile = k2_tile.to(gemm_dtype)
    v2_tile = v2_tile.to(gemm_dtype)

    dq = tl.zeros((BLOCK_SIZE_Q, HEAD_DIM), tl.float32)
    dk2 = tl.zeros((BLOCK_SIZE_KV2, HEAD_DIM), tl.float32)
    dv2 = tl.zeros((BLOCK_SIZE_KV2, HEAD_DIM), tl.float32)

    kv1_start = tl.maximum(0, q_start - w1)
    kv1_end = tl.minimum(seq_len, q_end)
    for kv1_idx in tl.range(kv1_start, kv1_end, num_stages=num_stages):
        k1_offs = kv1_idx * k1_stride_s + qkv_offs_h * k1_stride_h
        v1_offs = kv1_idx * v1_stride_s + qkv_offs_h * v1_stride_h
        k1_tile = tl.load(K1_ptr + k1_offs, mask=qkv_mask_h).to(
            compute_dtype
        )  # [HEAD_DIM]

        v1_tile = tl.load(V1_ptr + v1_offs, mask=qkv_mask_h).to(
            compute_dtype
        )  # [HEAD_DIM]

        qk1_s = q_tile * (k1_tile[None, :] * softmax_scale) # [Q, D]
        qk1_s = qk1_s.to(gemm_dtype)
        # k2[KV, Q] @ qk1_s.T[Q, D] => [KV2, Q]
        qkkT = tl.dot(k2_tile, qk1_s.T, out_dtype=tl.float32) # [KV2, Q]

        qkT_mask = kv2_mask_s[:, None] & q_mask_s[None, :] 
        kv1_local_mask = ((q_offs_s[None, :] - w1) < kv1_idx) & (
            kv1_idx <= q_offs_s[None, :]
        )  # [KV2, Q]
        kv2_local_mask = ((q_offs_s[None, :] - w2) < kv2_offs_s[:, None]) & (
            kv2_offs_s[:, None] <= q_offs_s[None, :]
        )  # [KV2, Q]
        local_mask = (
            kv1_local_mask & kv2_local_mask
        )  # [BLOCK_SIZE_KV, BLOCK_SIZE_Q]
        qkT_mask &= kv1_local_mask & kv2_local_mask

        pT = tl.exp(qkkT - m_tile[None, :]) # [KV2, Q]
        pT = tl.where(qkT_mask, pT, 0.0)

        qkkT = tl.where(local_mask, qkkT, -1.0e38)

        dOv1 = dO_tile * v1_tile[None, :] # [Q, D]
        dOv1 = dOv1.to(gemm_dtype)
        # pT[KV2, Q] @ dOv1[Q, D] => [KV2, D]
        dv2 += tl.dot(pT.to(gemm_dtype), dOv1, out_dtype=tl.float32) 
        
        # v2[KV2, D] @ dOv1.T[D, Q] => dpT[KV2, Q]
        dpT = tl.dot(v2_tile, dOv1.T, out_dtype=tl.float32)
        dsT = pT * (dpT - d_tile[None, :]) # [KV2, Q]
        dsT = tl.where(qkT_mask, dsT, 0.0)
        dsT = dsT.to(gemm_dtype) # [KV2, Q]

        # dsT[KV2, Q] @ qk1[Q, D] => dk2[KV2, D]
        dk2 += tl.dot(dsT, qk1_s, out_dtype=tl.float32)

        k1k2 = k1_tile[None, :] * k2_tile # [KV2, D]
        k1k2 = k1k2.to(gemm_dtype)        

        dq += tl.dot(dsT.T, k1k2) # * softmax scale at the end.

    # End. update derivatives.
    if IS_SECOND_PASS:
        #load, add.
        prev_dk2 = tl.load(dK2_ptr + kv2_offs, kv2_mask)
        prev_dv2 = tl.load(dV2_ptr + kv2_offs, kv2_mask)
        dk2 += prev_dk2
        dv2 += prev_dv2
    
    dq *= softmax_scale
    tl.store(dK2_ptr + kv2_offs, dk2, kv2_mask)
    tl.store(dV2_ptr + kv2_offs, dv2, kv2_mask)
    tl.store(dQ_ptr + q_offs, dq, q_mask)
\end{lstlisting}
\end{document}

%% file: math_commands.tex
\usepackage{amsmath,amsfonts,bm,amsthm}

\def\eqref#1{equation~\ref{#1}}

\def\1{\bm{1}}

\def\vk{{\bm{k}}}

\def\vq{{\bm{q}}}

\def\vv{{\bm{v}}}

\DeclareMathAlphabet{\mathsfit}{\encodingdefault}{\sfdefault}{m}{sl}
\SetMathAlphabet{\mathsfit}{bold}{\encodingdefault}{\sfdefault}{bx}{n}

\newcommand{\R}{\mathbb{R}}

\newtheorem{theorem}{Theorem}[section]

\theoremstyle{definition}

\theoremstyle{remark}

\newcommand{\inabs}[1]{\lvert #1 \rvert}
\newcommand{\inangle}[1]{\langle #1 \rangle}

\renewcommand{\vk}{\mathbf{k}}
\newcommand{\bigO}{\mathcal{O}}